\pdfoutput=1
\documentclass[11pt]{article}

\usepackage[]{emnlp2021}

\usepackage{times}
\usepackage{latexsym}

\usepackage[T1]{fontenc}
\usepackage[utf8]{inputenc}

\usepackage{microtype}
\usepackage{multirow}
\usepackage{graphicx}

\title{CSDS: A Fine-Grained Chinese Dataset for Customer Service Dialogue Summarization}

\author{Haitao Lin\textsuperscript{1,2}, Liqun Ma\textsuperscript{1,2}, Junnan Zhu\textsuperscript{1,2}, Lu Xiang\textsuperscript{1,2}, \\
{\bf Yu Zhou\textsuperscript{1,3}\thanks{\ \ Corresponding author.}, Jiajun Zhang\textsuperscript{1,2}, Chengqing Zong\textsuperscript{1,2}} \\
\textsuperscript{1} National Laboratory of Pattern Recognition, Institute of Automation, CAS, Beijing, China \\
\textsuperscript{2} School of Artificial Intelligence, University of Chinese Academy of Sciences, Beijing, China \\
\textsuperscript{3} Fanyu AI Laboratory, Zhongke Fanyu Technology Co., Ltd, Beijing, China\\
\texttt{$\left\{\right.$haitao.lin, junnan.zhu, lu.xiang, yzhou, jjzhang, } \\
\texttt{cqzong$\left .\right\}$@nlpr.ia.ac.cn, }
\texttt{maliqun2020@ia.ac.cn}
}

\begin{document}
\maketitle
\begin{abstract}
Dialogue summarization has drawn much attention recently. Especially in the customer service domain, agents could use dialogue summaries to help boost their works by quickly knowing customer's issues and service progress. These applications require summaries to contain the perspective of a single speaker and have a clear topic flow structure, while neither are available in existing datasets. Therefore, in this paper, we introduce a novel Chinese dataset for Customer Service Dialogue Summarization (CSDS). CSDS improves the abstractive summaries in two aspects: (1) In addition to the overall summary for the whole dialogue, role-oriented summaries are also provided to acquire different speakers' viewpoints. (2) All the summaries sum up each topic separately, thus containing the topic-level structure of the dialogue. We define tasks in CSDS as generating the overall summary and different role-oriented summaries for a given dialogue. Next, we compare various summarization methods on CSDS, and experiment results show that existing methods are prone to generate redundant and incoherent summaries. Besides, the performance becomes much worse when analyzing the performance on role-oriented summaries and topic structures. We hope that this study could benchmark Chinese dialogue summarization and benefit further studies.
\end{abstract}

\section{Introduction}

Text summarization aims to compress a long input text and generate a condensed summary \citep{zong2021text}. It can help people capture the gist of a long document quickly. Traditional summarization tasks mainly focus on news reports \citep{nallapati2016abstractive,narayan2018don,10.1007/978-3-319-73618-1_2}. However, as the communication tools become convenient, enormous information is presented in a conversational format, such as meeting records, daily chatting, and customer service logs. These dialogues usually cost more time to read since they are longer and have more complicated structures. Thus, summarizing information from dialogue becomes essential in practical use.  

Compared with news and documents, dialogues have two main unique features. First, dialogues have multiple speakers, and each of them has different viewpoints. In some cases, we only focus on the main viewpoint of one participant. A role-oriented summary will help achieve this goal. Second, a dialogue often has multiple topics, and each topic concerns a different issue. Since the summary expresses the primary information of the dialogue briefly and clearly, it should contain the topic flow of the dialogue by summing up each topic separately and be organized in a more structural format \citep{Zou_Zhao_Kang_Lin_Peng_Jiang_Sun_Zhang_Huang_Liu_2021}. Specifically, both features are rather crucial for practical applications in the customer service domain:  (1) The user-oriented summary could reflect the frequency of users' issues, and the agent-oriented summary could help check the quality of agents' services. (2) Topic-based structural summary could help an agent clearly acquire the user's problems and previous service progress, figuring out the solved and unsolved problems. Besides, role-oriented and structural summaries are also valuable for other dialogue domains such as debating and court trials.

\begin{figure*}[tp] 
    \centering  
    \includegraphics[scale=0.63]{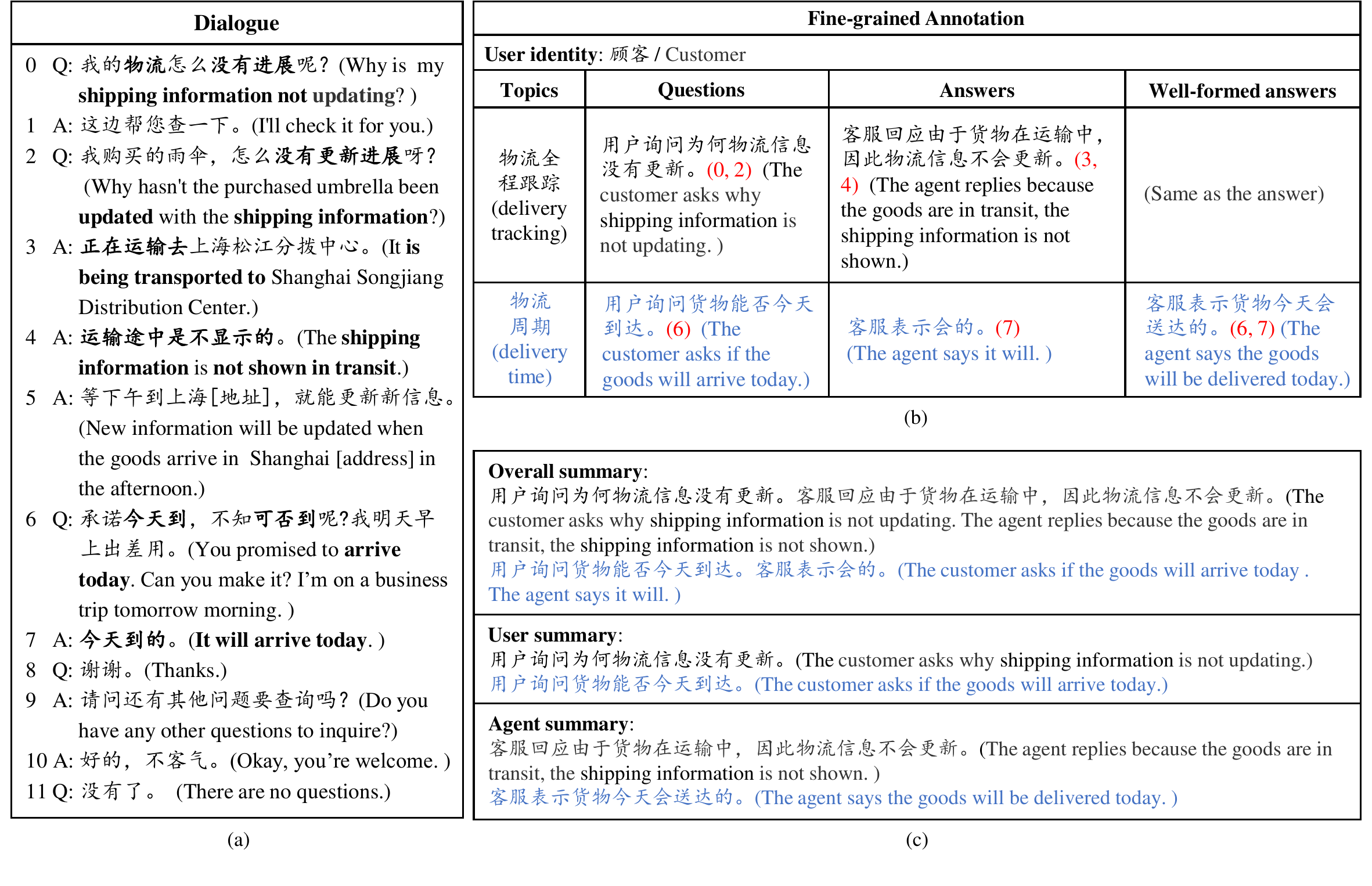}
    \caption{An annotation example for CSDS dataset. The annotation contents in black and blue represent summaries of different topics. We translate all Chinese texts into English for illustration. Red numbers are key utterance indexes, and bold texts represent key information.} 
    \label{fig:data-example}
\end{figure*}

Although several dialogue summarization datasets have been proposed recently \citep{mccowan2005ami,gliwa2019samsum,Zou_Lin_Zhao_Kang_Jiang_Sun_Zhang_Huang_Liu_2021,Zou_Zhao_Kang_Lin_Peng_Jiang_Sun_Zhang_Huang_Liu_2021}, none of them adds dialogue features (\textit{i.e.}, different speakers' roles or topic structure) in summaries, limiting the application of these datasets. Therefore, we aim to construct a fine-grained Chinese dataset for Customer Service domain Dialogue Summarization (CSDS). 

To achieve this goal, we employ Question-Answer (QA) pairs as the annotation format since it is the basic granularity in a customer service dialogue. We ask annotators to divide the dialogue into segments based on the topic and summarize each segment into a QA pair. Each pair consists of a user's question summary and an agent's answer summary. Next, we modify these QA pairs and recombine them into three types of summaries, \textit{i.e.,} the overall summary for the whole dialogue, the user summary for user's viewpoints, and the agent summary for agent's viewpoints.  Besides, we also annotate key utterances in the dialogue that provide critical information for summaries. These utterances could be considered as extractive summaries. An annotation example is shown in Figure \ref{fig:data-example}. Benefiting from the annotation, our labeled summaries are \textbf{role-oriented} (containing different speakers' main viewpoints) and \textbf{topic-based structural} (summarizing different topics separately). 

In all, we obtain more than 10,000 fine-grained dialogue annotations and more than 30,000 dialogue-summary pairs in three types. Next, we provide benchmark methods, including both extractive and abstractive methods, for CSDS. Comparing the automatic and human evaluation results, we find that existing methods are prone to make mistakes, including: (1) The questions and answers are often mismatched in the overall summary. (2) Generated summaries usually contain unnecessary and repeated texts. Meanwhile, there are some unique challenges in CSDS: (1) Some essential information in the agent summary is usually missing when it needs to integrate messages from users' utterances.  (2) Existing methods could not summarize separate topics correctly, especially when the number of topics increases in the dialogue. We also provide a specific evaluation metric based on QA pair matching for comparison and analysis. Additionally, we observe that most abstractive methods can boost their performance by using our annotated key utterance indexes, which can guide the further study of dialogue summarization methods.

The contributions of this paper are two-fold: (1) To the best of our knowledge, we are the first to construct a dialogue summarization dataset focusing on dialogue features. Each dialogue has summaries for different roles, and each summary sums up different topics separately\footnote{We make the dataset and related code available on https://github.com/xiaolinAndy/CSDS.}. (2) We elaborately compare and analyze the results of different summarization methods on our tasks. We conclude critical difficulties of tasks in CSDS and 
helpful information for boosting the performance of existing methods. We hope that our work could benchmark Chinese dialogue summarization and promote the development of this area. 

\section{Related Work}

Most of the widely-used summarization datasets belong to document summarization. Early datasets are provided by some evaluation tasks, such as DUC \citep{dang2005overview} and TAC \citep{dang2008overview}. Recently, there exist various types of document summarization datasets such as news reports \citep{nallapati2016abstractive,narayan2018don}, wiki passages \cite{liu2018generating}, and scientific papers \cite{kang2018dataset}.

Different from document summarization, dialogue summarization aims to summarize a conversation into a narrative text. \citet{mccowan2005ami,janin2003icsi} provide two datasets for dialogue summarization at the earliest, and the task of their data is to summarize a meeting transcript into a few short sentences. \citet{zhong-etal-2021-qmsum} incorporate them and propose a query-based summarization task. 
\citet{gliwa2019samsum} propose an English daily conversation summarization dataset on fictitious dialogues, providing a new daily chatting scenario. Other datasets provided by \citet{rameshkumar2020storytelling}, \citet{duan2019legal} and \citet{zhu-etal-2021-mediasum} also show the potential of dialogue summarization in other scenarios. As for Chinese dialogue summarization datasets, \citet{song2020summarizing} construct a medical dialogue summarization dataset, where most of the summaries are extractive and relatively easy to generate. Almost all the above datasets only provide an overall summary for each dialogue without further annotations.

In the customer service domain, \citet{Zou_Zhao_Kang_Lin_Peng_Jiang_Sun_Zhang_Huang_Liu_2021} provide a related dialogue summarization dataset, which is the most similar to our work. However, their dataset only contains an overall summary from the agent's perspective for each dialogue. Besides, their publicly available data are difficult to analyze since all the sentences are given by word indexes. On the contrary, our dataset is readable and has more fine-grained annotations such as role-oriented summaries and topic-based structural summaries, both valuable and convenient for further research in this area.

\section{Dataset Construction}
\label{sec:dataset}

\subsection{Data Collection}
We built our dataset based on JDDC \citep{chen2020jddc}.  
JDDC is a large-scale multi-turn Chinese dialogue dataset containing conversations about pre-sales and after-sales topics between users and agents in a real-world e-commerce scenario.

First, we selected some dialogues in JDDC which satisfy the following requirements. 
(1) We tend to select dialogues with long turns to ensure sufficient information and difficulty for summarization.
(2) Topics are various and evenly distributed. This requirement is achieved by controlling the distribution of query intents among the selected dialogues.
(3) All dialogues should be semantically complete (not truncated from a long conversation).
More details are given in Appendix A.

Next, we annotated our CSDS data based on these dialogues. 
Note that all private information in the CSDS has been anonymized, consistent with JDDC. More details are given in the Ethical Considerations Section.

\subsection{Fine-grained Annotation Format}

For each dialogue, our annotation contains three components, \textit{i.e.}, user identity, question-answer (QA) pairs, and key utterance indexes.

\paragraph{User identity}
As shown in Figure \ref{fig:data-example} (a), speakers in each dialogue are represented as \textit{Q} and \textit{A}. \textit{A} stands for the customer service agent, while \textit{Q} has three different identities (customer, seller, and deliveryman). Besides, the user identity is unique for every single dialogue. Thus, we annotated the identity based on the dialogue and replace \textit{Q} with it in the dialogue. This process keeps the expression consistent in the dialogue and the summary.

\paragraph{QA pairs}

First, we would like to find out a universal format for the customer service dialogue summary. Thus we took a pilot experiment by letting three well-educated annotators summarize the same 50 dialogues. We limited the summary to be less than 100 words and counted the formats of their summaries. We give the result in appendix B. The statistics reveal that most annotated summaries consist of several segments. Each segment focuses on a single topic in a QA pair format (The user raises a problem, and the agent gives a solution).

Based on the result of the pilot experiment, we believe that the QA pair is an appropriate format for dialogue summarization. To obtain a dialogue summary, we divided each dialogue into several segments according to different discussion topics and summarized the content of each topic independently into a QA pair. We present an example in Figure \ref{fig:data-example} (b). Note that the ``question'' and ``answer'' are not the same as defined for dialogue acts \citep{bunt-etal-2012-iso}, but at a more abstractive level. The ``question'' summarizes the problem or question raised by the user, while the ``answer'' summarizes the problem-solving process provided by the agent, thus also including questions raised by agents. Besides, each QA pair was annotated with a topic label according to the intent classes of dialogue utterances provided in JDDC dataset. We describe more details in appendix C.

In addition to summaries, this form of annotations can be extended to construct pseudo-QA pairs from dialogue for training models of downstream tasks, such as QA systems.

\paragraph{Key utterance indexes}
Key utterances are dialogue utterances that provide critical information for summaries. We asked the annotators to annotate the indexes of key utterances that reflect the important information in the dialogue. These annotations could be regarded as extractive summaries labels. An example is shown in Figure \ref{fig:data-example} (b) with red font. 

\subsection{Summary Format}
\label{ssec:summ-form}

Based on the annotations above, we could obtain three different summaries for each dialogue, including an overall summary and two role-oriented summaries (user summary and agent summary). We give an example in Figure \ref{fig:data-example} (c).

\paragraph{Overall summary}
The overall summary condenses the main information of the whole dialogue. We concatenated annotated QA pairs in sequence to obtain the overall summary. Each QA pair represents the summary for a single topic in the dialogue, and the order reflects the topic flow.

\paragraph{User summary}
The user summary only focuses on the user's main viewpoints and often includes the user's problems, questions, and explanations. Benefiting from the QA pair annotation, we only need to concatenate all the question parts in QA pairs and considered it as the user summary.

\paragraph{Agent summary}
The agent summary only focuses on the agent's responses, consisting of solutions to problems, answers to questions, and inquiries from the agent. 
However, different from questions, the agent's answers in the QA pair might have ellipsis, \textit{e.g.}, \textit{``Yes''} or \textit{``I will''}. These answers are only meaningful and readable with their related questions. An example is given in Figure \ref{fig:data-example} (b). The answer \textit{``The agent says it will''} is incomplete without the corresponding question. Thus we elaborately completed these answers by adding necessary contexts to ensure that they can be well understood alone. Finally, we concatenated these well-formed answers to obtain the agent summary.

\subsection{Human Annotation Process}
\label{ssec:anno-process}
We hired 44 undergraduate students to annotate the dialogues mentioned above. Each annotator first labeled the user identity in the dialogue.
Next, the annotator summarized QA pairs and annotated key utterance indexes.
We demanded that questions in QA pairs be refined from users' core questions in an objective form, and answers contain the process of solving related questions and the final results.
Any trivial information, like greeting or appreciation, should be omitted in QA pairs.

Lastly, we followed the process in Section \ref{ssec:summ-form} to obtain three types of summaries from the annotated data.
For agent summaries, we filtered out answers difficult to be understood without their related questions. As explained in \ref{ssec:summ-form}, we asked four annotators to complete these answers into well-formed ones and obtained agent summaries by concatenating them. More details are in Appendix D. We present an example of the whole annotation process and the acquired summaries in Figure \ref{fig:data-example}.

\begin{table*}[tp]
\centering
\small
\begin{tabular}{cccccccc}
\hline

\textbf{Dataset} & \textbf{Lang.} & \textbf{\# Dialogues}  & \textbf{\begin{tabular}[c]{@{}c@{}}Dialogue\\ Source\end{tabular}}             & \textbf{\begin{tabular}[c]{@{}c@{}}Role\\ Sum.\end{tabular}} & \textbf{\begin{tabular}[c]{@{}c@{}}Sum.\\ Type\end{tabular}} & \textbf{\begin{tabular}[c]{@{}c@{}}Readability\end{tabular}} &
\textbf{\begin{tabular}[c]{@{}c@{}}Topic \\ Structure\end{tabular}} \\ \hline
AMI              & EN             & 97 / 20 / 20                & meeting record                                                                 & No                                                            & Abs.                                                         & Yes            & No      \\ \hline
SAMSum           & EN             & 14,732 / 818 / 819         & fictitious chat                                                                & No                                                            & Abs.                                                         & Yes         & No        \\ \hline
CRD3            & EN             & 26,232 / 3,470 / 4,541       & \begin{tabular}[c]{@{}c@{}}role-playing\\ game transcripts\end{tabular}        & No                                                            & Abs.                                                         & Yes          & No        \\ \hline
\cite{song2020summarizing}              & ZH             & 35,987 / 8,996            & \begin{tabular}[c]{@{}c@{}}online medical\\ conversation\end{tabular}          & Yes                                                           & Ext.                                                         & Yes        & No          \\ \hline
\cite{Zou_Zhao_Kang_Lin_Peng_Jiang_Sun_Zhang_Huang_Liu_2021}              & ZH             & 17,189 / 820 / 851         & \begin{tabular}[c]{@{}c@{}}real-world\\ customer service\end{tabular}          & No                                                            & Abs.                                                         & No             & No      \\ \hline
PLD              & EN             & 4,500 / 977              & \begin{tabular}[c]{@{}c@{}}court debate\\ records\end{tabular}                 & No                                                            & Ext.                                                         & No            & No       \\ \hline
\textbf{CSDS}    & \textbf{ZH}    & \textbf{9,101 / 800 / 800} & \textbf{\begin{tabular}[c]{@{}c@{}}real-world\\ customer service\end{tabular}} & \textbf{Yes}                                                  & \textbf{\begin{tabular}[c]{@{}c@{}}Abs. \\\& Ext.\end{tabular}}                                        & \textbf{Yes}    & \textbf{Yes}     \\ \hline
\end{tabular}
\caption{\label{tab:CSDS-compare} The comparison of different dialogue summarization datasets.
\textit{\#Dialogues} represents the size of train/validation/test set for each dataset.
\textit{Readability} represents whether the contents in the dataset are readable.
\textit{Topic Structure} represents whether the dialogue summary summarizes each topic in the dialogue separately.}
\end{table*}

\begin{table}[]
\small
\begin{tabular}{c|cccc}
\hline
\textbf{} & \textbf{\# Dial.} & \textbf{\begin{tabular}[c]{@{}c@{}}Dial.\\Length\end{tabular}} & \textbf{\# Turns} & \textbf{\# QA Pairs} \\ \hline
Train     & 9,101              & 401.08         & 26.00           & 1.98             \\
Dev.      & \ \ 800               & 396.34         & 25.90           & 1.99             \\
Test      & \ \ 800               & 387.10         & 25.11           & 1.90             \\ \hline
\end{tabular}
%\begin{tabular}[c]{@{}c@{}}Dial.\\Length\end{tabular}
~\\

\begin{tabular}{c|cc}
\hline
\textbf{}     & \textbf{\begin{tabular}[c]{@{}c@{}}Sum. Length\\ (Min/Avg/Max)\end{tabular}} & \textbf{\begin{tabular}[c]{@{}c@{}}Compression Ratio\\ (Min/Avg/Max)\end{tabular}} \\ \hline
Overall Sum. & 11/83.21/475       & 0.012/0.220/0.875             \\
User Sum.     & \ \ 5/37.28/239        & 0.006/0.099/0.489             \\
Agent Sum.    & \ \ 7/48.08/266        & 0.006/0.127/0.608             \\ \hline
\end{tabular}
\caption{\label{tab:CSDS-stat} Some statistical information of CSDS. \textit{Dial. Length} and \textit{Sum. Length} represent the number of characters in dialogues and summaries. \textit{Compression Ratio} is calculated as \textit{Sum. Length} / \textit{Dial. Length.}}
\end{table}

\subsection{Quality Control}
\label{ssec:quality}
To ensure that each annotator can finish the task with high quality, we set up a pre-annotation test. We first let the annotators read our instructions thoroughly and asked them to annotate five test samples. Two expert examiners on this task checked whether the annotation satisfied the following four criteria: (1) The extracted QA pairs contain all core information in the dialogue. (2) There is no redundant information appearing in QA pairs. (3) All the QA pairs are fluent and easy to understand. (4) The key utterance indexes and user identity are labeled correctly. The annotators that met all the criteria could continue to label the formal data. After labeling the formal data, two examiners sampled 10\% of the annotators' data and checked them carefully.  If the acceptability ratio was lower than 90\%, the corresponding annotators were asked to revise their annotation.
The loop ended until the acceptability ratio reached 90\%.

Since summarization is a relatively subjective task, it is impossible to control different annotators generating the same summaries. We control the consistency of annotated summaries by employing a more objective examination process as mentioned above. To ensure the reliability of the above four criteria, we did the inter-annotator agreement study between examiners. Two examiners evaluated the quality of the same 100 samples by judging whether each annotation meets the four criteria. The kappa scores on the four rules were 0.51, 0.61, 0.55, 0.65. Since the evaluation of NLG tasks is a more challenging task and the inter-rater agreement cannot be very high \citep{amidei-etal-2018-rethinking}. These reasonable inter-annotator agreements show the reasonability of our quality control criteria and the credibility of the annotated summaries for further research.

\subsection{Dataset Statistics and Comparison}
\label{ssec:stat}

We compare our dataset with other dialogue summarization datasets in Table \ref{tab:CSDS-compare} and some statistics of CSDS is shown in Table \ref{tab:CSDS-stat}. We summarize the advantages of CSDS in the following three aspects: 

%\noindent\textbf{Multiple Roles' Perspectives}
\paragraph{Multiple Roles' Perspectives}
Since CSDS have summaries for different roles, they can reflect the main viewpoints of the dialogue through different perspectives. It also raises a question on how to express the main point of one speaker while maintaining complete semantic information.

\paragraph{Topic Structure}
Our summaries are split by different topics, thus maintaining the topic flow in the dialogue. This kind of format could reflect the dialogue content more clearly and straightforwardly. Meanwhile, we could also evaluate the summary quality in a topic-level granularity.

\paragraph{Key Utterance Annotation}
CSDS annotates the key utterance indexes in each dialogue and maps them with the related summaries. They could be used as extractive summary references and additional training signals to help boost summarization performance.

\section{Task and Experiment Setup}

\subsection{Task Definition}

The input of the task is a dialogue with multiple turns. Each utterance is labeled with its speaker role (\textit{e.g.}, user or agent) and the specific user identity (\textit{e.g.}, customer or seller). The task is to generate three different kinds of summaries as explained in Section \ref{ssec:summ-form}, including the overall summary, user summary, and agent summary. Each summary should consist of several segments, and each segment represents the summary for a topic in the dialogue. Besides, models could use the annotated key utterance indexes as additional supervised signals during the training process. 

However, existing methods for dialogue summarization are not specified for role-oriented summaries or guaranteed to generate a topic-based structural summary, which are the specific features in CSDS. Thus, we want to figure out what performance existing methods could reach by training to generate different kinds of summaries separately and relax the structural requirements\footnote{We do not give these methods any topic information in both training and test phase. However, this information may help different methods generate summaries with higher quality. We will leave it to future work.}.

\subsection{Summarization Models}
\label{ssec:models}

In this section, we will introduce some widely-used extractive and abstractive summarization models on dialogue summarization. We also enhance some of the models using our special annotations. The extractive methods include:

\textbf{LONGEST}: As the longer utterances in the dialogue may contain more useful information, we sort the utterances by their lengths and extract the top $k$ longest utterances as the summary. Number $k$ is decided by the maximum summary length limit.  

\textbf{LexPageRank} \citep{erkan2004lexpagerank}: This method ranks dialogue utterances by PageRank algorithm and extracts utterances in order until the length of the summary reaches the limit.

\textbf{SummaRuNNer} \citep{nallapati2017summarunner}: A supervised extractive summarization method by scoring each utterance using RNN. Here, we use the key utterance indexes as extractive labels.

\textbf{BERTExt} \citep{liu2019text}: This method scores each utterance in dialogue by finetuning on the pretrained BERT \citep{devlin2019bert} model. Extractive labels are the same as SummaRuNNer.

We also implement some abstractive methods:

\textbf{PGN} \citep{see2017get}: An RNN-based seq2seq model using source word copy mechanism and attention coverage mechanism. 

\textbf{Fast-RL} \citep{chen2018fast}: This method first extracts important sentences and then compresses them into summaries. The whole model is at last jointly trained by reinforcement learning.

\textbf{BERTAbs} \citep{liu2019text}: It uses pretrained BERT as the encoder and a transformer-based network as the decoder to summarize.

\textbf{TDS+SATM} \citep{Zou_Zhao_Kang_Lin_Peng_Jiang_Sun_Zhang_Huang_Liu_2021}: It is similar to Fast-RL but uses BERT and transformer structure as the extractive model and abstractive model. Besides, it also introduces a topic model to enhance summary generation.

For all abstractive methods containing the extractive process, such as Fast-RL and TDS+SATM, we also use the annotated \textbf{key utterance indexes} as supervised signals only in the training process. We name them as \textbf{Fast-RL*} and \textbf{TDS+SATM*}. 

Besides, all extractive methods are restricted to generate summaries less than a limited length, which is 84 for the overall summary, 38 for the user summary, and 49 for the agent summary. They are set according to the average length of reference summaries. More experimental settings are given in Appendix E.

\subsection{Evaluation Metrics}
\label{ssec:metrics}

We employ five widely used automatic metrics to evaluate the above methods. The automatic metrics\footnote{We use the F1 score variant of all the metrics if multiple variants exist.} include:

\textbf{ROUGE-based methods} \citep{lin2002manual}: Widely used metrics by measuring the overlap of n-grams between two texts. Here we choose \textbf{ROUGE-2} and \textbf{ROUGE-L} for comparison.

\textbf{BLEU} \citep{papineni2002bleu}: Another n-gram overlap metric by considering up to 4-grams. 

\textbf{BERTScore} \citep{zhang2019bertscore}: It measures the word overlap between two texts according to contextual BERT embeddings.

\textbf{MoverScore} \citep{zhao2019moverscore}: It measures the semantic distance between two texts according to pretrained embeddings. Here we use BERT embedding as well.

\begin{table*}
\centering
\small
\begin{tabular}{lccccc}
\hline
\textbf{Methods}          & \textbf{ROUGE-2} & \textbf{ROUGE-L} & \textbf{BLEU} & \textbf{BERTScore} & \textbf{MoverScore} \\ \hline
LONGEST       &    15.52/20.26/13.84       & 22.18/30.53/21.63       & 11.19/13.14/9.94    & 63.61/67.92/62.89     & 12.38/16.46/10.71       \\
LexPageRank   &    19.43/19.29/16.56       & 26.86/30.59/25.92       & 13.48/14.14/12.65   & 66.60/67.23/65.27     & 15.01/13.94/12.26       \\
SummaRunner   &    27.99/26.46/25.26       & 37.91/40.16/36.36       & 21.60/19.35/20.69   & 71.77/72.16/70.94     & 24.10/22.16/20.41       \\
BERTExt       &    27.51/21.58/23.05       & 32.99/32.59/29.48       & 21.59/14.91/17.39   & 71.24/68.01/67.59     & 22.69/16.06/14.59      \\ \hline
PGN           &  39.19/37.05/35.19       &  \textbf{47.94}/48.57/45.11   & 32.31/29.64/28.29     & 78.40/78.67/76.15    & 28.58/26.65/25.17      \\
Fast-RL       &  \textbf{41.39}/40.43/\textbf{37.59}  &    47.07/51.49/\textbf{46.30}     & \textbf{33.04}/33.39/\textbf{30.44}  & 79.57/80.29/\textbf{77.72}   & 29.78/28.55/\textbf{27.18}       \\
Fast-RL*     &  41.24/\textbf{41.68}/37.38       &    47.27/\textbf{52.83}/45.55     & 32.94/\textbf{33.53}/30.11     & \textbf{79.76}/\textbf{81.06}/77.52      & \textbf{30.12}/\textbf{29.95}/26.89     \\
BERTAbs      & 37.03/35.15/33.20        &    45.30/46.22/42.89     & 24.59/27.76/24.73     &  78.45/78.76/76.35     & 27.00/24.20/23.67     \\
TDS+SATM     &  33.19/34.29/32.76       &  42.43/47.19/43.62       & 20.24/22.44/23.51      &  76.84/78.25/76.07      & 24.29/24.95/24.09   \\
TDS+SATM*    & 35.33/34.91/30.56        & 44.38/47.94/41.89        & 24.68/22.94/20.94     & 77.79/78.60/75.19     & 26.16/25.04/22.56     \\ \hline
\end{tabular}
\caption{\label{tab:auto-summary} The automatic metric results for overall summary and role-oriented summaries, each block has three values, representing overall summary/user summary/agent summary from left to right.}
\end{table*}

As for human evaluation metrics, we try to evaluate the quality of summaries at a fine-grained topic level. First, we split the ground truth summaries and the summaries generated by models into different topic segments\footnote{The ground truth summary is split according to the QA pair annotations, and the generated summary is split by humans.}. Then we evaluate the summary quality for each segment in the following three aspects: informativeness, non-redundancy, fluency\footnote{We found that this kind of evaluation is more objective to reflect the quality of summaries since evaluating a summary for a single topic is easier to achieve agreements.}. These three aspects are frequently used in the summarization community \citep{zhu-etal-2019-ncls, 10.1162/tacl_a_00373}, and we also refer to some researches on NLG evaluation \citep{howcroft-etal-2020-twenty, belz-etal-2020-disentangling}. Each aspect is scored on a three-point scale, 0 for the worst, 1 for the medium, and 2 for the best. Three aspects are defined as:

\textbf{Informativeness}: How much key information of the ground truth summary does the generated summary correctly cover?

\textbf{Non-redundancy}: Does the generated summary contain repeated, meaningless or unnecessary information?

\textbf{Fluency}: Is the generated summary formal, well-formed, grammatically correct? 

In addition to the above three aspects, we propose a new metric named \textbf{Matching Rate}. It represents the matching rate of questions and answers in the overall summaries. Moreover, it can reflect the semantic coherency of summaries since unmatched QA pairs can lead to huge incoherence between sentences.

\begin{table*}
\centering
\small
\begin{tabular}{lcccc}
\hline
\textbf{Methods}          & \textbf{Informativeness} & \textbf{Non-redundancy} & \textbf{Fluency} & \textbf{Matching Rate} \\ \hline
PGN              & 1.18/1.08/1.20        & 0.91/0.93/\textbf{1.01}        & 1.41/1.47/1.64      &  \textbf{0.73}             \\
Fast-RL          & \textbf{1.27}/\textbf{1.16}/\textbf{1.25}      & 0.94/\textbf{1.11}/0.93         & 1.49/\textbf{1.58}/\textbf{1.69}       &  0.57           \\
BERTAbs          & 0.73/0.72/0.74        & \textbf{1.01}/0.77/0.99         & \textbf{1.57}/1.50/1.58        &  0.62           \\
TDS+SATM*          & 0.90/0.78/0.79        & 0.99/1.03/0.95         & 1.37/1.46/1.66        & 0.57             \\ \hline
\end{tabular}
\caption{\label{tab:human-summary} The human evaluation results for the overall summary and role-oriented summary, each block has three values and represents the same as in Table \ref{tab:auto-summary}. For the first three metrics, 0 stands for the worst and 2 for the best. Matching Rate ranges from 0 to 1 and is only available for the overall summary.}
\end{table*}

\section{Experimental Results}

\subsection{Automatic Evaluation Results}

First, we present automatic evaluation metric results of different models in Table \ref{tab:auto-summary}. In general, we observe that abstractive methods perform better than extractive methods with a large margin. Among extractive methods, SummaRunner achieves the best results, indicating the effectiveness of supervised utterance index labels. As for abstractive methods, Fast-RL and Fast-RL* perform best on almost all metrics except for ROUGE-L of the overall summary, where the PGN method obtains a better result. Transformer-based methods perform worse mainly because of relatively small data size \citep{joshi-etal-2020-dr}. It is worth noticing that enhanced methods (Fast-RL*, TDS+SATM*) are usually better than their original version on the overall summary and the user summary. This highlights the effect of the key utterance indexes even just used as supervised signals, as it can reflect which utterance is more critical for summarization. 

By comparing with the same model in different tasks, we find that the agent summary scores are much lower than the overall summary and user summary in most metrics. It demonstrates that generating agent summaries is more difficult than the other two types of summaries since it needs to focus on what the agent says and incorporate some necessary information from the user.

\subsection{Human Evaluation Results}

Next, we choose the outputs of some summarization methods and let humans evaluate them in the metrics we defined in Section \ref{ssec:metrics}. We choose four relatively well-performed methods and randomly sample 50 dialogues from the test set. We recruit three well-educated volunteers to evaluate the three different types of summaries generated from four methods. We also run the inter-annotator agreement study on three volunteers' scores, and the kappa score is 0.52 on average. The result is shown in Table \ref{tab:human-summary}. 

Obviously, all methods perform poorly on non-redundancy, where most of the scores are lower than one. Besides, they also achieve low informativeness scores. These results prove that although some methods can reach high automatic metric scores, the generated summaries still contain much useless information and miss some essential content. Moreover, we find that nearly 30 percent of overall summaries have unmatched questions and answers through the matching rate. It demonstrates that these methods could not guarantee to generate a semantically coherent summary.

\section{Dataset Difficulties}

In this section, we want to analyze the difficulties of CSDS further. According to the fine-grained features in CSDS and the challenges mentioned in Section \ref{ssec:stat}, we raise the following two questions. (1) Compared with the overall summary, what are the difficulties for the role-oriented summary?  (2) Could existing methods generate summaries with the correct topic structure?

\paragraph{Integrating messages from other roles for the role-oriented summary is difficult.}

Compared with the overall summary, the role-oriented summary focuses on a single role's utterances. It needs to integrate messages from other roles to make the summary understandable, especially for agent summary. To analyze whether existing methods could learn to integrate different roles' information for agent summary, we compare the summary quality of samples which need to be integrated and those that do not need separately. Note that in Section \ref{ssec:summ-form}, we complete some agent summaries to make them well-understood without contexts. Thus, these \textbf{completed summaries} are considered as requiring information integration and others as not requiring integration.

As shown in Table \ref{tab:role-summary-diff}, all three models obtain much lower ROUGE scores\footnote{We also conduct other metrics on this experiment, and they show similar results with ROUGE scores. More details are given in Appendix F.} on agent summary for samples that need to integrate than other samples, indicating the insufficient ability to provide a semantically intact agent summary. However, the overall summary results do not show the same trend, proving that this gap is only caused by agent summaries. More specific models, such as jointly learning to generate different types of summaries, need to be studied for this task.

\begin{table}
\centering
\small
\begin{tabular}{llll}
\hline
\textbf{Methods}           & \textbf{Summ.} & \textbf{\begin{tabular}[c]{@{}c@{}}ROUGE-L\\Type A/B\end{tabular}} & \textbf{\begin{tabular}[c]{@{}c@{}}BERTScore\\Type A/B\end{tabular}} \\ \hline
\multirow{2}{*}{PGN}       & overall          & \textbf{54.69}/52.95    &   \textbf{80.23}/78.82                 \\
                           & agent            & 49.77/\textbf{56.46}    &   76.74/\textbf{78.23}                 \\ \hline
\multirow{2}{*}{Fast-RL}   & overall          & \textbf{56.00}/52.79    &   \textbf{81.97}/79.90                 \\
                           & agent            & 48.92/\textbf{54.38}    &   77.62/\textbf{78.65}                 \\ \hline
\multirow{2}{*}{TDS+SATM*} & overall          & 42.94/\textbf{43.66}    &   \textbf{77.71}/77.15                 \\
                           & agent            & 34.87/\textbf{43.46}    &   72.65/\textbf{74.73}                 \\ \hline
\end{tabular}
\caption{\label{tab:role-summary-diff} The performance of some methods on different types of samples. Type A stands for agent summaries that need to be integrated, and Type B stands for those that do not. Note that all the metrics here are recall scores.}
\end{table}

\begin{table}
\centering
\small
\begin{tabular}{llll}
\hline
\textbf{Methods}          & \textbf{Precision} & \textbf{Recall} & \textbf{F1} \\ \hline
PGN              &   \textbf{0.188}      &  0.210       &  \textbf{0.199}            \\
Fast-RL          &   0.146      &  \textbf{0.221}       &  0.176                \\
TDS+SATM*          &  0.174       &  0.130       &  0.149          \\ \hline
\end{tabular}
\caption{\label{tab:qa-pair-num} The ratio of correctly summarized QA pair for some baseline methods.}
\end{table}

\begin{figure}[htbp] 
    \centering  
    \includegraphics[scale=0.4]{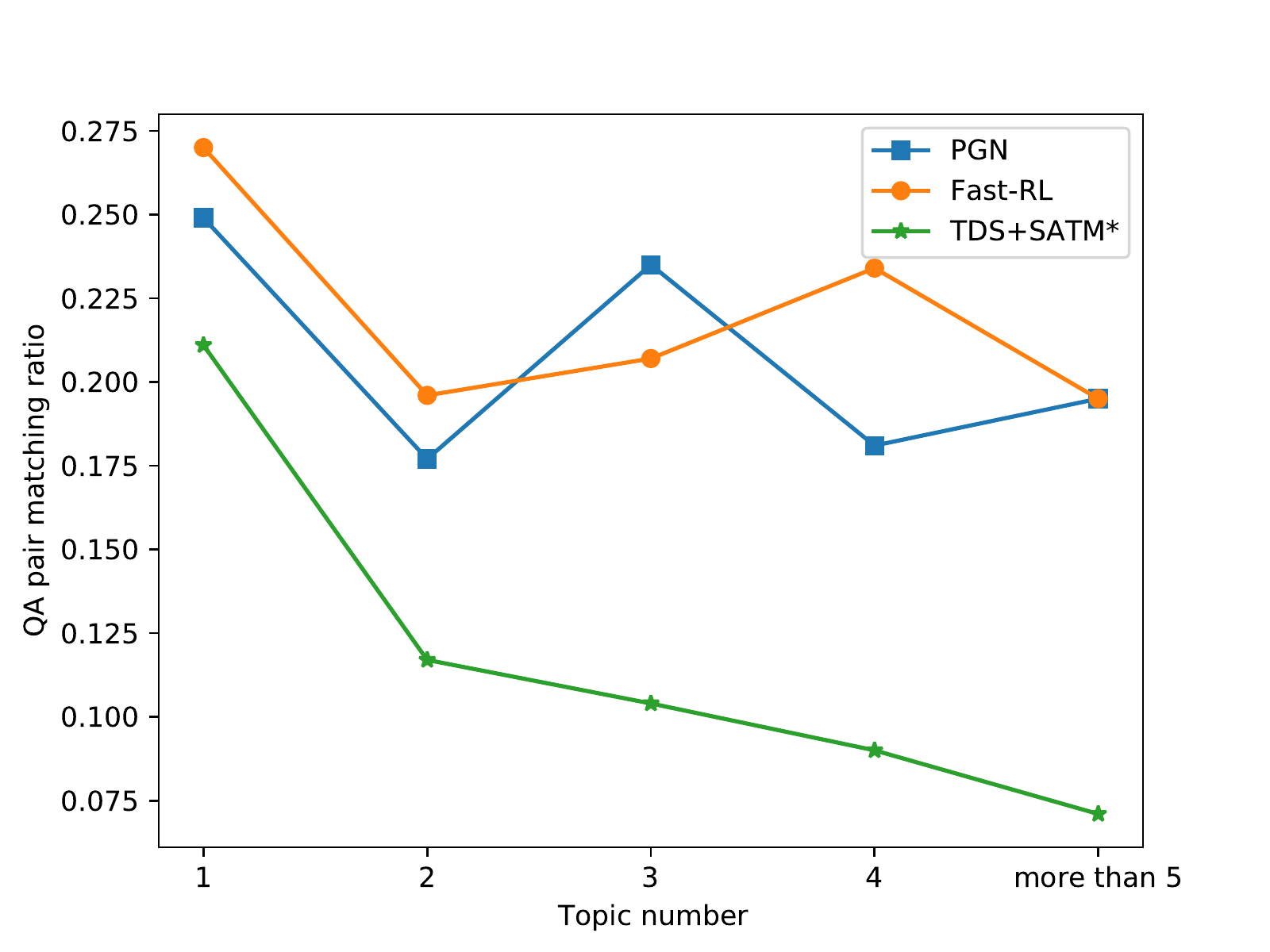}
    \caption{The relationship between the recall score of QA pair matching ratio and the number of QA pairs in the reference summary, which is also the topic number.} 
    \label{fig:qa-ratio}
\end{figure}

\paragraph{Generating summaries with the correct topic structure is difficult.}

Traditional summarization models treat the dialogue as a whole and do not specifically consider the structure of generated summaries. Thus we wonder whether these methods could generate summaries with correct topic structure as references. We analyze the overall summary results and consider each QA pair in the summary as a whole, judging whether each QA pair is contained by the results of different models.

We provide a ROUGE-L-based greedy match algorithm to calculate the number of correct QA pairs (More details are given in Appendix G). We calculate the precision, recall, and F1 scores of correctly matched QA pair ratio and present them in Table \ref{tab:qa-pair-num}. The result shows that the best model can only match around 20\% of QA pairs in the summary. Besides, more than 80\% of redundant QA pairs exist in the generated summary since the best precision score is only 0.19. Both show the poor ability of existing methods to separate and summarize QA pairs, which is not shown by calculating the ROUGE score in general.

We also analyze how the QA pair matching ratio changes with the number of QA pairs in the reference summary and present it in Figure \ref{fig:qa-ratio}. The similar trends for three different methods indicate that it is harder to separate QA pairs for different topics when the number of QA pairs increases in reference summaries. Therefore, how to summarize dialogues with the correct topic structure needs further study. Besides, our evaluation algorithm can also be used as a standard metric to compare the performance of different methods on CSDS.

At last, we present some examples of mistakes made by existing methods on CSDS in appendix H. From these cases, we conclude that more elaborate methods are needed for dialogue summarization.

\section{Conclusion}

In this paper, we introduce a novel customer service dialogue summarization dataset named CSDS. This dataset is fine-grained in two folds. Summaries for different roles are provided, and summaries contain topic structure information, which could help promote research and applications in this area. We also do elaborate experiments on CSDS and draw some instructive conclusions on method performance and dataset difficulties. In the future, we hope that we can propose new models for CSDS to summarize different roles or summarize the content for a specific topic. More suitable automatic evaluation metrics, especially for comparing the topic structure in the summary, are also worth studying.

\section*{Acknowledgements}

First, We thank anonymous reviewers for helpful suggestions. Second, we thank all the annotators and volunteers for constructing the dataset and making the human evaluation. This work was supported by the National Key R\&D Program of China under Grant No. 2020AAA0108600.

\section*{Ethical Considerations}

\subsection*{Privacy and Licensing Issues}

CSDS is a dataset constructed from JDDC dataset \citep{chen2020jddc}. As mentioned in their paper, they anonymized private information in the dataset, such as replacing all numbers with a special token <NUM> and all order IDs with <ORDER-ID>. Meanwhile, no personal information of customers and agents in JDDC is released. Our annotations are merely based on the information in JDDC. Thus there is no private information in CSDS as well.

\subsection*{Annotator Information and Compensation}

We estimated that a skillful annotator needs 3 to 5 minutes to finish an annotation for each dialogue. Therefore, we paid annotators 2 yuan (\$0.3) for each dialogue. We also paid the same bonus for the pilot experiment. The compensation for completing agent summaries and summary evaluation is 1 yuan (\$0.15) per dialogue since they are easier than writing summaries. We added the compensation rules into the recruitment document, and all the annotators accepted it as a reasonable standard before they agreed to annotate data.

We also present some demographic information about the annotators in Table \ref{tab:demo}. Note that although the annotators come from different regions in China, we ensured that they are proficient in mandarin and demanded they write summaries in mandarin. Thus the dialect has less influence on the summary quality.

\begin{table}[htbp]
\centering
\small
\begin{tabular}{ll}
\hline
Demographic Information & Value                             \\ \hline
Amount                  & 44                                \\
Gender (Male / Female)  & 20 / 24                           \\
Age                     & {[}18, 25{]}                      \\
Native Language         & All Chinese                       \\
Education Background    & Undergraduate or graduate \\ \hline
\end{tabular}
\caption{\label{tab:demo} The demographic information of annotators.}
\end{table}

\subsection*{Dataset Characteristics and Generalizability}

CSDS is a dialogue summarization dataset used to summarize a customer service dialogue. We describe the details of our data filtering process in Appendix A. The demographic information about users and agents in the dialogue is unavailable. To ensure the generalizability of our dataset, we did an IAA study as mentioned in \ref{ssec:quality}. The moderate agreements prove that our annotated summaries are reliable and have good generalizability.

% Entries for the entire Anthology, followed by custom entries
\bibliography{anthology,custom}

\begin{thebibliography}{36}
\expandafter\ifx\csname natexlab\endcsname\relax\def\natexlab#1{#1}\fi

\bibitem[{Amidei et~al.(2018)Amidei, Piwek, and
  Willis}]{amidei-etal-2018-rethinking}
Jacopo Amidei, Paul Piwek, and Alistair Willis. 2018.
\newblock \href {https://www.aclweb.org/anthology/C18-1281} {Rethinking the
  agreement in human evaluation tasks}.
\newblock In \emph{Proceedings of the 27th International Conference on
  Computational Linguistics}, pages 3318--3329, Santa Fe, New Mexico, USA.
  Association for Computational Linguistics.

\bibitem[{Belz et~al.(2020)Belz, Mille, and
  Howcroft}]{belz-etal-2020-disentangling}
Anya Belz, Simon Mille, and David~M. Howcroft. 2020.
\newblock \href {https://www.aclweb.org/anthology/2020.inlg-1.24}
  {Disentangling the properties of human evaluation methods: A classification
  system to support comparability, meta-evaluation and reproducibility
  testing}.
\newblock In \emph{Proceedings of the 13th International Conference on Natural
  Language Generation}, pages 183--194, Dublin, Ireland. Association for
  Computational Linguistics.

\bibitem[{Bunt et~al.(2012)Bunt, Alexandersson, Choe, Fang, Hasida, Petukhova,
  Popescu-Belis, and Traum}]{bunt-etal-2012-iso}
Harry Bunt, Jan Alexandersson, Jae-Woong Choe, Alex~Chengyu Fang, Koiti Hasida,
  Volha Petukhova, Andrei Popescu-Belis, and David Traum. 2012.
\newblock \href
  {http://www.lrec-conf.org/proceedings/lrec2012/pdf/530_Paper.pdf} {{ISO}
  24617-2: A semantically-based standard for dialogue annotation}.
\newblock In \emph{Proceedings of the Eighth International Conference on
  Language Resources and Evaluation ({LREC}'12)}, pages 430--437, Istanbul,
  Turkey. European Language Resources Association (ELRA).

\bibitem[{Chen et~al.(2020)Chen, Liu, Shen, Yuan, Zhou, Wu, He, and
  Zhou}]{chen2020jddc}
Meng Chen, Ruixue Liu, Lei Shen, Shaozu Yuan, Jingyan Zhou, Youzheng Wu,
  Xiaodong He, and Bowen Zhou. 2020.
\newblock \href {https://www.aclweb.org/anthology/2020.lrec-1.58} {The {JDDC}
  corpus: A large-scale multi-turn {C}hinese dialogue dataset for {E}-commerce
  customer service}.
\newblock In \emph{Proceedings of the 12th Language Resources and Evaluation
  Conference}, pages 459--466, Marseille, France. European Language Resources
  Association.

\bibitem[{Chen and Bansal(2018)}]{chen2018fast}
Yen-Chun Chen and Mohit Bansal. 2018.
\newblock \href {https://doi.org/10.18653/v1/P18-1063} {Fast abstractive
  summarization with reinforce-selected sentence rewriting}.
\newblock In \emph{Proceedings of the 56th Annual Meeting of the Association
  for Computational Linguistics (Volume 1: Long Papers)}, pages 675--686,
  Melbourne, Australia. Association for Computational Linguistics.

\bibitem[{Dang(2005)}]{dang2005overview}
Hoa~Trang Dang. 2005.
\newblock Overview of duc 2005.
\newblock In \emph{Proceedings of the document understanding conference},
  volume 2005, pages 1--12.

\bibitem[{Dang and Owczarzak(2008)}]{dang2008overview}
Hoa~Trang Dang and Karolina Owczarzak. 2008.
\newblock Overview of the tac 2008 update summarization task.
\newblock In \emph{TAC}.

\bibitem[{Devlin et~al.(2019)Devlin, Chang, Lee, and
  Toutanova}]{devlin2019bert}
Jacob Devlin, Ming-Wei Chang, Kenton Lee, and Kristina Toutanova. 2019.
\newblock \href {https://doi.org/10.18653/v1/N19-1423} {{BERT}: Pre-training of
  deep bidirectional transformers for language understanding}.
\newblock In \emph{Proceedings of the 2019 Conference of the North {A}merican
  Chapter of the Association for Computational Linguistics: Human Language
  Technologies, Volume 1 (Long and Short Papers)}, pages 4171--4186,
  Minneapolis, Minnesota. Association for Computational Linguistics.

\bibitem[{Duan et~al.(2019)Duan, Zhang, Yuan, Zhou, Liu, Wang, Wang, Zhang,
  Sun, and Wu}]{duan2019legal}
Xinyu Duan, Yating Zhang, Lin Yuan, Xin Zhou, Xiaozhong Liu, Tianyi Wang,
  Ruocheng Wang, Qiong Zhang, Changlong Sun, and Fei Wu. 2019.
\newblock \href {https://doi.org/10.1145/3357384.3357940} {Legal summarization
  for multi-role debate dialogue via controversy focus mining and multi-task
  learning}.
\newblock In \emph{Proceedings of the 28th {ACM} International Conference on
  Information and Knowledge Management, {CIKM} 2019, Beijing, China, November
  3-7, 2019}, pages 1361--1370. {ACM}.

\bibitem[{Erkan and Radev(2004)}]{erkan2004lexpagerank}
G{\"u}ne{\c{s}} Erkan and Dragomir~R. Radev. 2004.
\newblock \href {https://www.aclweb.org/anthology/W04-3247} {{L}ex{P}age{R}ank:
  Prestige in multi-document text summarization}.
\newblock In \emph{Proceedings of the 2004 Conference on Empirical Methods in
  Natural Language Processing}, pages 365--371, Barcelona, Spain. Association
  for Computational Linguistics.

\bibitem[{Fabbri et~al.(2021)Fabbri, Kryściński, McCann, Xiong, Socher, and
  Radev}]{10.1162/tacl_a_00373}
Alexander~R. Fabbri, Wojciech Kryściński, Bryan McCann, Caiming Xiong,
  Richard Socher, and Dragomir Radev. 2021.
\newblock \href {https://doi.org/10.1162/tacl_a_00373} {{SummEval:
  Re-evaluating Summarization Evaluation}}.
\newblock \emph{Transactions of the Association for Computational Linguistics},
  9:391--409.

\bibitem[{Gliwa et~al.(2019)Gliwa, Mochol, Biesek, and Wawer}]{gliwa2019samsum}
Bogdan Gliwa, Iwona Mochol, Maciej Biesek, and Aleksander Wawer. 2019.
\newblock \href {https://doi.org/10.18653/v1/D19-5409} {{SAMS}um corpus: A
  human-annotated dialogue dataset for abstractive summarization}.
\newblock In \emph{Proceedings of the 2nd Workshop on New Frontiers in
  Summarization}, pages 70--79, Hong Kong, China. Association for Computational
  Linguistics.

\bibitem[{Howcroft et~al.(2020)Howcroft, Belz, Clinciu, Gkatzia, Hasan,
  Mahamood, Mille, van Miltenburg, Santhanam, and
  Rieser}]{howcroft-etal-2020-twenty}
David~M. Howcroft, Anya Belz, Miruna-Adriana Clinciu, Dimitra Gkatzia, Sadid~A.
  Hasan, Saad Mahamood, Simon Mille, Emiel van Miltenburg, Sashank Santhanam,
  and Verena Rieser. 2020.
\newblock \href {https://www.aclweb.org/anthology/2020.inlg-1.23} {Twenty years
  of confusion in human evaluation: {NLG} needs evaluation sheets and
  standardised definitions}.
\newblock In \emph{Proceedings of the 13th International Conference on Natural
  Language Generation}, pages 169--182, Dublin, Ireland. Association for
  Computational Linguistics.

\bibitem[{Janin et~al.(2003)Janin, Baron, Edwards, Ellis, Gelbart, Morgan,
  Peskin, Pfau, Shriberg, Stolcke et~al.}]{janin2003icsi}
Adam Janin, Don Baron, Jane Edwards, Dan Ellis, David Gelbart, Nelson Morgan,
  Barbara Peskin, Thilo Pfau, Elizabeth Shriberg, Andreas Stolcke, et~al. 2003.
\newblock The icsi meeting corpus.
\newblock In \emph{2003 IEEE International Conference on Acoustics, Speech, and
  Signal Processing, 2003. Proceedings.(ICASSP'03).}, volume~1, pages I--I.
  IEEE.

\bibitem[{Joshi et~al.(2020)Joshi, Katariya, Amatriain, and
  Kannan}]{joshi-etal-2020-dr}
Anirudh Joshi, Namit Katariya, Xavier Amatriain, and Anitha Kannan. 2020.
\newblock \href {https://doi.org/10.18653/v1/2020.findings-emnlp.335} {Dr.
  summarize: Global summarization of medical dialogue by exploiting local
  structures.}
\newblock In \emph{Findings of the Association for Computational Linguistics:
  EMNLP 2020}, pages 3755--3763, Online. Association for Computational
  Linguistics.

\bibitem[{Kang et~al.(2018)Kang, Ammar, Dalvi, van Zuylen, Kohlmeier, Hovy, and
  Schwartz}]{kang2018dataset}
Dongyeop Kang, Waleed Ammar, Bhavana Dalvi, Madeleine van Zuylen, Sebastian
  Kohlmeier, Eduard Hovy, and Roy Schwartz. 2018.
\newblock \href {https://doi.org/10.18653/v1/N18-1149} {A dataset of peer
  reviews ({P}eer{R}ead): Collection, insights and {NLP} applications}.
\newblock In \emph{Proceedings of the 2018 Conference of the North {A}merican
  Chapter of the Association for Computational Linguistics: Human Language
  Technologies, Volume 1 (Long Papers)}, pages 1647--1661, New Orleans,
  Louisiana. Association for Computational Linguistics.

\bibitem[{Lin and Hovy(2002)}]{lin2002manual}
Chin-Yew Lin and Eduard Hovy. 2002.
\newblock \href {https://doi.org/10.3115/1118162.1118168} {Manual and automatic
  evaluation of summaries}.
\newblock In \emph{Proceedings of the {ACL}-02 Workshop on Automatic
  Summarization}, pages 45--51, Phildadelphia, Pennsylvania, USA. Association
  for Computational Linguistics.

\bibitem[{Liu et~al.(2018)Liu, Saleh, Pot, Goodrich, Sepassi, Kaiser, and
  Shazeer}]{liu2018generating}
Peter~J. Liu, Mohammad Saleh, Etienne Pot, Ben Goodrich, Ryan Sepassi, Lukasz
  Kaiser, and Noam Shazeer. 2018.
\newblock \href {https://openreview.net/forum?id=Hyg0vbWC-} {Generating
  wikipedia by summarizing long sequences}.
\newblock In \emph{6th International Conference on Learning Representations,
  {ICLR} 2018, Vancouver, BC, Canada, April 30 - May 3, 2018, Conference Track
  Proceedings}. OpenReview.net.

\bibitem[{Liu and Lapata(2019)}]{liu2019text}
Yang Liu and Mirella Lapata. 2019.
\newblock \href {https://doi.org/10.18653/v1/D19-1387} {Text summarization with
  pretrained encoders}.
\newblock In \emph{Proceedings of the 2019 Conference on Empirical Methods in
  Natural Language Processing and the 9th International Joint Conference on
  Natural Language Processing (EMNLP-IJCNLP)}, pages 3730--3740, Hong Kong,
  China. Association for Computational Linguistics.

\bibitem[{McCowan et~al.(2005)McCowan, Carletta, Kraaij, Ashby, Bourban, Flynn,
  Guillemot, Hain, Kadlec, Karaiskos et~al.}]{mccowan2005ami}
I~McCowan, J~Carletta, W~Kraaij, S~Ashby, S~Bourban, M~Flynn, M~Guillemot,
  T~Hain, J~Kadlec, V~Karaiskos, et~al. 2005.
\newblock The ami meeting corpus.
\newblock In \emph{Proceedings of Measuring Behavior 2005, 5th International
  Conference on Methods and Techniques in Behavioral Research}, pages 137--140.
  Noldus Information Technology.

\bibitem[{Nallapati et~al.(2017)Nallapati, Zhai, and
  Zhou}]{nallapati2017summarunner}
Ramesh Nallapati, Feifei Zhai, and Bowen Zhou. 2017.
\newblock \href {http://aaai.org/ocs/index.php/AAAI/AAAI17/paper/view/14636}
  {Summarunner: {A} recurrent neural network based sequence model for
  extractive summarization of documents}.
\newblock In \emph{Proceedings of the Thirty-First {AAAI} Conference on
  Artificial Intelligence, February 4-9, 2017, San Francisco, California,
  {USA}}, pages 3075--3081. {AAAI} Press.

\bibitem[{Nallapati et~al.(2016)Nallapati, Zhou, dos Santos,
  G{\.{u}}l{\c{c}}ehre, and Xiang}]{nallapati2016abstractive}
Ramesh Nallapati, Bowen Zhou, Cicero dos Santos, {\c{C}}a{\u{g}}lar
  G{\.{u}}l{\c{c}}ehre, and Bing Xiang. 2016.
\newblock \href {https://doi.org/10.18653/v1/K16-1028} {Abstractive text
  summarization using sequence-to-sequence {RNN}s and beyond}.
\newblock In \emph{Proceedings of The 20th {SIGNLL} Conference on Computational
  Natural Language Learning}, pages 280--290, Berlin, Germany. Association for
  Computational Linguistics.

\bibitem[{Narayan et~al.(2018)Narayan, Cohen, and Lapata}]{narayan2018don}
Shashi Narayan, Shay~B. Cohen, and Mirella Lapata. 2018.
\newblock \href {https://doi.org/10.18653/v1/D18-1206} {Don{'}t give me the
  details, just the summary! topic-aware convolutional neural networks for
  extreme summarization}.
\newblock In \emph{Proceedings of the 2018 Conference on Empirical Methods in
  Natural Language Processing}, pages 1797--1807, Brussels, Belgium.
  Association for Computational Linguistics.

\bibitem[{Papineni et~al.(2002)Papineni, Roukos, Ward, and
  Zhu}]{papineni2002bleu}
Kishore Papineni, Salim Roukos, Todd Ward, and Wei-Jing Zhu. 2002.
\newblock \href {https://doi.org/10.3115/1073083.1073135} {{B}leu: a method for
  automatic evaluation of machine translation}.
\newblock In \emph{Proceedings of the 40th Annual Meeting of the Association
  for Computational Linguistics}, pages 311--318, Philadelphia, Pennsylvania,
  USA. Association for Computational Linguistics.

\bibitem[{Rameshkumar and Bailey(2020)}]{rameshkumar2020storytelling}
Revanth Rameshkumar and Peter Bailey. 2020.
\newblock \href {https://doi.org/10.18653/v1/2020.acl-main.459} {Storytelling
  with dialogue: {A} {Critical Role Dungeons and Dragons Dataset}}.
\newblock In \emph{Proceedings of the 58th Annual Meeting of the Association
  for Computational Linguistics}, pages 5121--5134, Online. Association for
  Computational Linguistics.

\bibitem[{See et~al.(2017)See, Liu, and Manning}]{see2017get}
Abigail See, Peter~J. Liu, and Christopher~D. Manning. 2017.
\newblock \href {https://doi.org/10.18653/v1/P17-1099} {Get to the point:
  Summarization with pointer-generator networks}.
\newblock In \emph{Proceedings of the 55th Annual Meeting of the Association
  for Computational Linguistics (Volume 1: Long Papers)}, pages 1073--1083,
  Vancouver, Canada. Association for Computational Linguistics.

\bibitem[{Song et~al.(2020)Song, Tian, Wang, and Xia}]{song2020summarizing}
Yan Song, Yuanhe Tian, Nan Wang, and Fei Xia. 2020.
\newblock \href {https://doi.org/10.18653/v1/2020.coling-main.63} {Summarizing
  medical conversations via identifying important utterances}.
\newblock In \emph{Proceedings of the 28th International Conference on
  Computational Linguistics}, pages 717--729, Barcelona, Spain (Online).
  International Committee on Computational Linguistics.

\bibitem[{Zhang et~al.(2020)Zhang, Kishore, Wu, Weinberger, and
  Artzi}]{zhang2019bertscore}
Tianyi Zhang, Varsha Kishore, Felix Wu, Kilian~Q. Weinberger, and Yoav Artzi.
  2020.
\newblock \href {https://openreview.net/forum?id=SkeHuCVFDr} {Bertscore:
  Evaluating text generation with {BERT}}.
\newblock In \emph{8th International Conference on Learning Representations,
  {ICLR} 2020, Addis Ababa, Ethiopia, April 26-30, 2020}. OpenReview.net.

\bibitem[{Zhao et~al.(2019)Zhao, Peyrard, Liu, Gao, Meyer, and
  Eger}]{zhao2019moverscore}
Wei Zhao, Maxime Peyrard, Fei Liu, Yang Gao, Christian~M. Meyer, and Steffen
  Eger. 2019.
\newblock \href {https://doi.org/10.18653/v1/D19-1053} {{M}over{S}core: Text
  generation evaluating with contextualized embeddings and earth mover
  distance}.
\newblock In \emph{Proceedings of the 2019 Conference on Empirical Methods in
  Natural Language Processing and the 9th International Joint Conference on
  Natural Language Processing (EMNLP-IJCNLP)}, pages 563--578, Hong Kong,
  China. Association for Computational Linguistics.

\bibitem[{Zhong et~al.(2021)Zhong, Yin, Yu, Zaidi, Mutuma, Jha, Awadallah,
  Celikyilmaz, Liu, Qiu, and Radev}]{zhong-etal-2021-qmsum}
Ming Zhong, Da~Yin, Tao Yu, Ahmad Zaidi, Mutethia Mutuma, Rahul Jha,
  Ahmed~Hassan Awadallah, Asli Celikyilmaz, Yang Liu, Xipeng Qiu, and Dragomir
  Radev. 2021.
\newblock \href {https://doi.org/10.18653/v1/2021.naacl-main.472} {{QMS}um: A
  new benchmark for query-based multi-domain meeting summarization}.
\newblock In \emph{Proceedings of the 2021 Conference of the North American
  Chapter of the Association for Computational Linguistics: Human Language
  Technologies}, pages 5905--5921, Online. Association for Computational
  Linguistics.

\bibitem[{Zhu et~al.(2021)Zhu, Liu, Mei, and Zeng}]{zhu-etal-2021-mediasum}
Chenguang Zhu, Yang Liu, Jie Mei, and Michael Zeng. 2021.
\newblock \href {https://doi.org/10.18653/v1/2021.naacl-main.474}
  {{M}edia{S}um: A large-scale media interview dataset for dialogue
  summarization}.
\newblock In \emph{Proceedings of the 2021 Conference of the North American
  Chapter of the Association for Computational Linguistics: Human Language
  Technologies}, pages 5927--5934, Online. Association for Computational
  Linguistics.

\bibitem[{Zhu et~al.(2019)Zhu, Wang, Wang, Zhou, Zhang, Wang, and
  Zong}]{zhu-etal-2019-ncls}
Junnan Zhu, Qian Wang, Yining Wang, Yu~Zhou, Jiajun Zhang, Shaonan Wang, and
  Chengqing Zong. 2019.
\newblock \href {https://doi.org/10.18653/v1/D19-1302} {{NCLS}: Neural
  cross-lingual summarization}.
\newblock In \emph{Proceedings of the 2019 Conference on Empirical Methods in
  Natural Language Processing and the 9th International Joint Conference on
  Natural Language Processing (EMNLP-IJCNLP)}, pages 3054--3064, Hong Kong,
  China. Association for Computational Linguistics.

\bibitem[{Zhu et~al.(2018)Zhu, Zhou, Li, Zhang, Zhou, and
  Zong}]{10.1007/978-3-319-73618-1_2}
Junnan Zhu, Long Zhou, Haoran Li, Jiajun Zhang, Yu~Zhou, and Chengqing Zong.
  2018.
\newblock Augmenting neural sentence summarization through extractive
  summarization.
\newblock In \emph{Natural Language Processing and Chinese Computing}, pages
  16--28, Cham. Springer International Publishing.

\bibitem[{Zong et~al.(2021)Zong, Xia, and Zhang}]{zong2021text}
Chengqing Zong, Rui Xia, and Jiajun Zhang. 2021.
\newblock \emph{Text Data Mining}.
\newblock Springer.

\bibitem[{Zou et~al.(2021{\natexlab{a}})Zou, Lin, Zhao, Kang, Jiang, Sun,
  Zhang, Huang, and Liu}]{Zou_Lin_Zhao_Kang_Jiang_Sun_Zhang_Huang_Liu_2021}
Yicheng Zou, Jun Lin, Lujun Zhao, Yangyang Kang, Zhuoren Jiang, Changlong Sun,
  Qi~Zhang, Xuanjing Huang, and Xiaozhong Liu. 2021{\natexlab{a}}.
\newblock \href {https://ojs.aaai.org/index.php/AAAI/article/view/17724}
  {Unsupervised summarization for chat logs with topic-oriented ranking and
  context-aware auto-encoders}.
\newblock \emph{Proceedings of the AAAI Conference on Artificial Intelligence},
  35(16):14674--14682.

\bibitem[{Zou et~al.(2021{\natexlab{b}})Zou, Zhao, Kang, Lin, Peng, Jiang, Sun,
  Zhang, Huang, and
  Liu}]{Zou_Zhao_Kang_Lin_Peng_Jiang_Sun_Zhang_Huang_Liu_2021}
Yicheng Zou, Lujun Zhao, Yangyang Kang, Jun Lin, Minlong Peng, Zhuoren Jiang,
  Changlong Sun, Qi~Zhang, Xuanjing Huang, and Xiaozhong Liu.
  2021{\natexlab{b}}.
\newblock \href {https://ojs.aaai.org/index.php/AAAI/article/view/17723}
  {Topic-oriented spoken dialogue summarization for customer service with
  saliency-aware topic modeling}.
\newblock \emph{Proceedings of the AAAI Conference on Artificial Intelligence},
  35(16):14665--14673.

\end{thebibliography}
\bibliographystyle{acl_natbib}

\appendix

\section*{Appendix}

\section{Data Filtering Process}

We do not randomly select dialogues from JDDC but limit the selection in two aspects: length and topic intent. Thus we will describe how we concretely do to refine our selection process.

\paragraph{Length control}

The average number of turns for each dialogue in JDDC is around 20, and the modal number is 14. However, since we want to construct a summarization dataset, a longer dialogue could bring challenges to our task. Thus we try to increase the length of dialogue in CSDS while keeping the distribution smooth and reasonable. Specifically, during the sampling process, we set a probability for each dialogue $d_i$ as below:

\begin{equation}
    p_i = (\frac{len(d_i)}{\max\limits_i len(d_i)})^2
\end{equation}

\begin{figure}[htbp] 
    \centering  
    \includegraphics[scale=0.5]{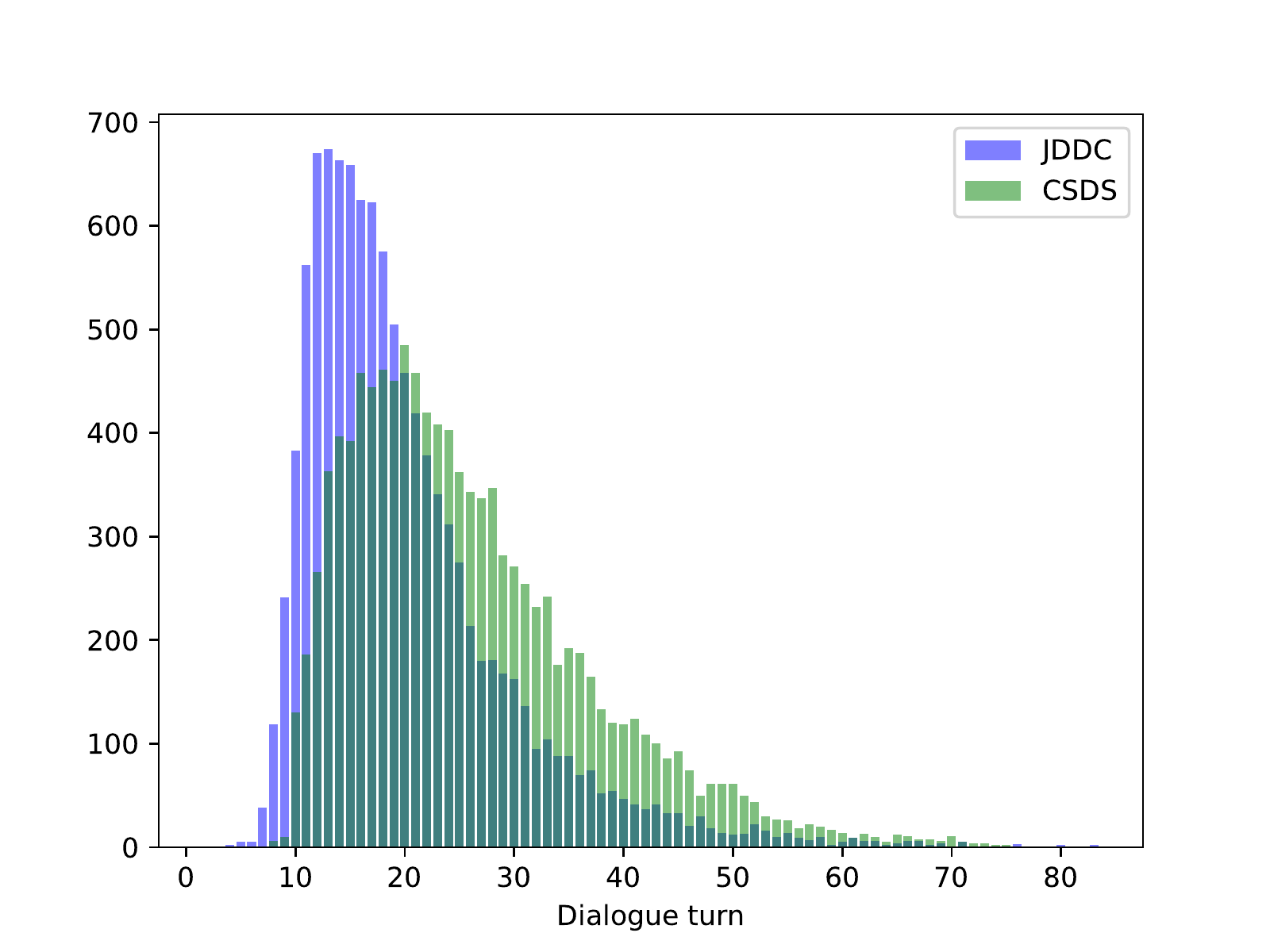}
    \caption{The distribution of dialogue turns in JDDC and CSDS.} 
    \label{fig:length}
\end{figure}

$len(d_i)$ represents the number of turns for dialogue $d_i$. It is obvious that as the dialogue becomes longer, its probability of being selected increases. We use this sampling strategy to obtain the dialogue in CSDS, and the average length increases to 26. The mode number is 20, and we compare the length distribution of JDDC and CSDS in Figure \ref{fig:length}. 

\paragraph{Topic intent control}

\begin{figure}[htbp] 
    \centering  
    \includegraphics[scale=0.5]{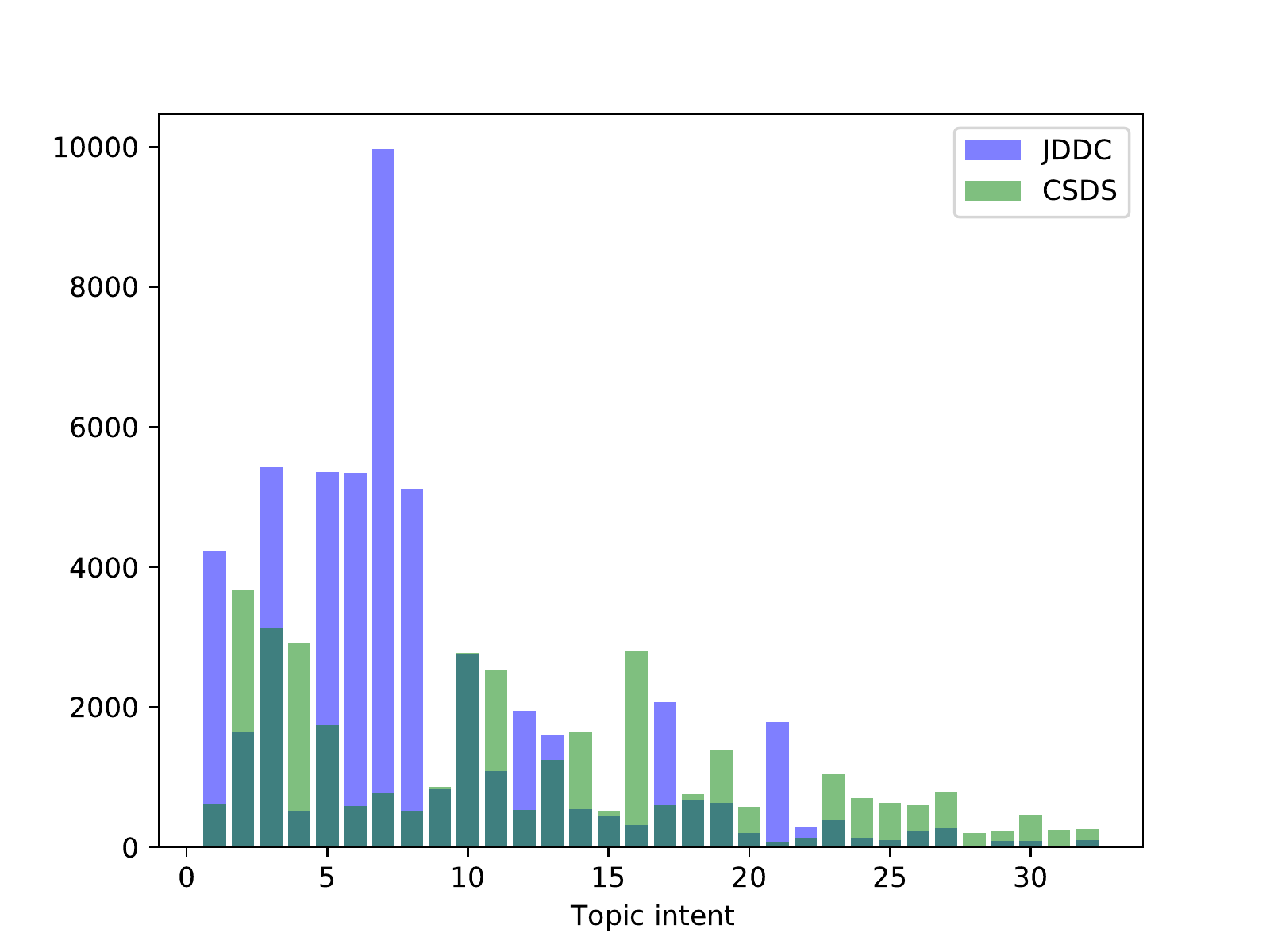}
    \caption{The distribution of topic intents in JDDC and CSDS. Note that each column represents a different topic intent.} 
    \label{fig:topic}
\end{figure}

In JDDC, each user utterance in the dialogue is labeled with an intent, indicating the topic information of the dialogue. However, the distribution of intents is also unbalanced, and most of the dialogues focus on the highly appeared intents such as return policy, shipping information, and invoice. To obtain various topics and increase the proportion of rare topics in CSDS, we try to balance the topic intent distribution using a particular strategy. We set a maximum number for each intent, which is 300 in practice. Note that there may exist multiple intents in one dialogue. Thus each time we select a dialogue, we add one to the counting for intents that appear in the dialogue. We will not select the dialogue if all of the intents in the dialogue have reached the maximum size. The comparison between the topic intent distribution of JDDC and CSDS is given in Figure \ref{fig:topic}. Obviously, CSDS flattens the topic distribution, and the ratio of some rare topics in JDDC is also increased.

\begin{figure*}[htbp] 
    \centering  
    \includegraphics[scale=0.71]{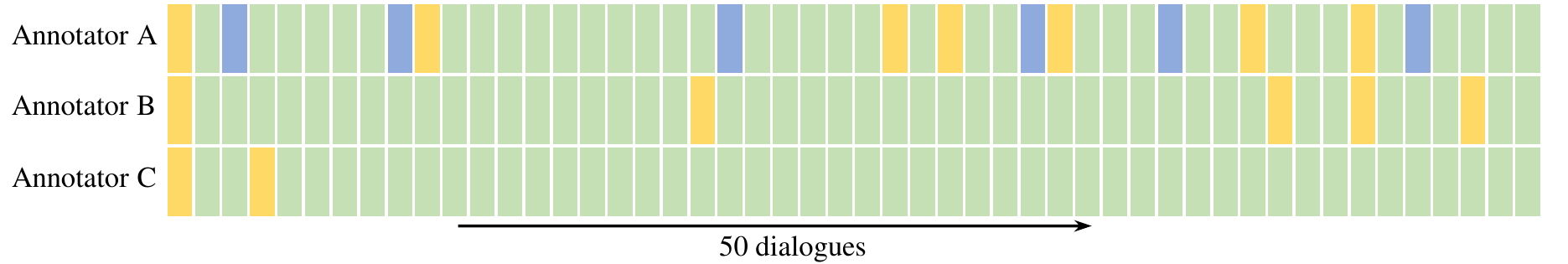}
    \caption{The summary form of 50 dialogues given by three annotators. The horizontal axis stands for different dialogue, and the vertical axis stands for different annotators. The green block indicates that the summary is in the QA pair form, the yellow one represents a non-QA pair summary, and the blue one stands for a summary similar to QA pairs.} 
    \label{fig:summ_form}
\end{figure*}

\section{QA Pair Priori Test}

In this section, we present the result of pilot experiment result for the summary format, as shown in Figure \ref{fig:summ_form}. Nearly 90 percent of the summaries are in a QA pair form without prior instruction. Meanwhile, although some of them (shown in blue) are not precisely in the QA pair form, they can also be easily converted to QA pairs. This pilot experiment proves the rationality of our proposed format.

However, there still exist some odd cases which are difficult to be summarized into QA pairs. In the formal annotation process, we asked the annotators to discard this kind of data to ensure the reliability of summaries.

\section{Topic Labels}

After obtaining annotated QA pairs and their key utterance indexes, we look up the intent labels of key utterances provided in JDDC dataset and sum up them in an intent set. If there is only one intent, we serve it as the topic label. If multiple intents exist, we confirm the topic label by choosing the intent with the highest frequency in the intent set or by manually checking if multiple intents have the same highest frequency. There are 289 different intents in JDDC, and 210 intents actually appear as topic labels in CSDS.

\section{Agent Summary Modifying Rules}

We give the rules to select agent summaries for completing as below:

\begin{enumerate}
    \item Length limit: We filter out the answers that are less than ten characters in each QA pair and consider it as a candidate to be completed. The reason is that short answers are more likely to omit words.
    \item Question and answer type limit: According to our observation, we find that the Yes / No questions are more prone to result in incomplete answers. Thus we use regular expression to filter out Yes / No questions if the QA pair meets the two following requirements: (1) The question summary includes words such as ``could'', ``can'', ``is'', ``are'', ``whether'', etc. (2) The answer summary contains words such as ``yes'', ``no'', ``need'', ``could'', etc.
\end{enumerate}

After the filtering process, we will let annotators decide whether the answer should be completed or not and make the completion if needed. In all, 26\% of the data is finally modified in the whole dataset. It proves the importance of completing agent summaries for CSDS.

\begin{table*}[htbp]
\centering
\small
\begin{tabular}{lllllll}
\hline
\textbf{Methods}           & \textbf{Summ.} & \textbf{\begin{tabular}[c]{@{}c@{}}ROUGE-2\\Type A/B\end{tabular}} & \textbf{\begin{tabular}[c]{@{}c@{}}ROUGE-L\\Type A/B\end{tabular}} & \textbf{\begin{tabular}[c]{@{}c@{}}BLEU\\Type A/B\end{tabular}} & \textbf{\begin{tabular}[c]{@{}c@{}}BERTScore\\Type A/B\end{tabular}} & \textbf{\begin{tabular}[c]{@{}c@{}}MoverScore\\Type A/B\end{tabular}} \\ \hline
\multirow{2}{*}{PGN}       & overall          & \textbf{45.26}/42.71    & \textbf{54.69}/52.95    & \textbf{29.43}/26.22   &   \textbf{80.23}/78.82                            &  28.46/\textbf{28.61}            \\
                           & agent            & 38.64/\textbf{43.31}    & 49.77/\textbf{56.46}    & 22.91/\textbf{23.60}   &   76.74/\textbf{78.23} &  \textbf{25.17}/\textbf{25.17}                 \\ \hline
\multirow{2}{*}{Fast-RL}   & overall          & \textbf{50.79}/45.83    & \textbf{56.00}/52.79    & \textbf{32.22}/27.11   &   \textbf{81.97}/79.90                            &  30.02/\textbf{30.16}   \\
                           & agent            & 40.96/\textbf{43.52}    & 48.92/\textbf{54.38}    & 24.72/\textbf{24.77}   &   77.62/\textbf{78.65} &  26.81/\textbf{26.90}                \\ \hline
\multirow{2}{*}{TDS+SATM*} & overall          & \textbf{34.63}/34.42    & 42.94/\textbf{43.66}    & \textbf{22.70}/22.05   &   \textbf{77.71}/77.15                            &  26.01/\textbf{26.23}  \\
                           & agent            & 25.28/\textbf{31.42}    & 34.87/\textbf{43.46}    & 14.31/\textbf{17.71}   &   72.65/\textbf{74.73} &  22.03/\textbf{22.08}                \\ \hline
\end{tabular}
\caption{\label{tab:app-diff-role} The performance of some methods on different types of samples. Type A stands for agent summaries that need to be integrated and Type B stands for those that do not. Note that all the metrics here are recall scores.}
\end{table*}

\section{Experimental Settings}

First, we will introduce the basic settings for all the models.

\begin{enumerate}
    \item We use the word-level granularity for all the models without BERT\footnote{It shows better performance than character-level in our prior experiments.}. The word segmentation for dialogue is provided in CSDS, and we use jieba\footnote{https://github.com/fxsjy/jieba/} to segment words in the summaries. Since BERT-based models process Chinese texts on the character level, we do not change the segmentation methods for these models.
    \item For all the models without BERT, we use pretrained Chinese word vectors provided by Tencent\footnote{https://ai.tencent.com/ailab/nlp/en/embedding.html}. The vocabulary size is 10000. While for BERT-based models, we use Chinese-BERT-wwm\footnote{https://github.com/ymcui/Chinese-BERT-wwm} pretrained embeddings.
    \item We add the speaker role information (user identity or agent) to the front of each utterance in the dialogue, ensuring that the input contains the speaker's information for different turns.
    \item The input dialogue limit is 500 words and 1000 characters, and the output summary limit is 100 words and 200 characters.
    \item For all abstractive methods, we use beam search to generate summaries, and the beam size is 5.
    \item The length limits for extractive methods are calculated according to the average summary length in the training set. We also try to use the average compression rate and find that the fixed length performs better.
\end{enumerate}

Then, there are some specific settings for every single method. All the parameters that are not mentioned are kept the same with the default settings in the open-source code.

\paragraph{PGN:}

Since PGN needs a single input for all the dialogue contexts, we concatenate all the utterances together and add a special token ``$<$EOU$>$'' to segment each utterance. The maximum training epoch is 30, and we finetune the model with the coverage mechanism for another 10 epochs. 

\paragraph{Fast-RL:}

We concatenate the same speaker's continuous utterances to obtain a long utterance as each selected utterance in Fast-RL is summarized into a sentence in the summary. The maximum number of utterances is 60, and the maximum number of words in each utterance is 100.

\paragraph{Fast-RL*:}

We use the annotated key utterance indexes to obtain the extractive labels by assigning the most similar utterance\footnote{It is calculated by ROUGE-L.} for each summary sentence from the corresponding key utterances.

\paragraph{BERTAbs:}

The original parameter for the learning rate is not suitable for our dataset. Therefore, we set the learning rate of BERT as 0.002 and decoder as 0.02. The maximum step is 4000, and all the settings in BERTExt are the same as those in BERTAbs.

\paragraph{TDS+SATM*:}

We merge all the key utterance indexes as the supervised signal for its extractive model.

All the experiments are run on a single NVIDIA TITAN Xp and the total time cost of all the methods is around a week. We use files2rouge\footnote{https://github.com/pltrdy/files2rouge} to calculate ROUGE scores, nltk\footnote{http://www.nltk.org/} to calculate BLEU, official python packages to calculate BERTScore\footnote{https://github.com/Tiiiger/bert\_score} and MoverScore\footnote{https://github.com/AIPHES/emnlp19-moverscore}. For ROUGE scores, all the Chinese characters are transferred into vocabulary ids for calculation.

We select the best hyper-parameters by choosing the best performance on the validation set with the minimum cross-entropy loss. The best hyper-parameters are given in the ``run.sh'' code for each model (The hyper-parameters not mentioned are consistent with the default settings in the model source code).

\section{Full Results of Role-oriented Summaries}

In Table \ref{tab:app-diff-role}, we compare the results of samples that need integration and those that do not need on all automatic evaluation methods. ROUGE-based metrics and BERTScore are their recall variant since we focus on whether the information from the other speaker is contained in the summary. Since BLEU and MoverScore do not have recall variants, we use the available result instead. Almost all metrics show the same trend that the performance of agent summaries on samples that needs integration is significantly lower than that on other samples.

\section{QA Pair Matching Algorithm}

First, we divide each generated summary into several QA pairs by considering contiguous sentences started by user and agent as a QA pair. Next, we try to match each QA pair in the reference with the QA pair in the generated summary using the ROUGE-L F1 score. We set the threshold to be 0.6\footnote{This setting has the highest classification accuracy in our experimental test.} and treat it as a match if the ROUGE-L F1 is higher than the threshold.

We define $QA_m$ as the number of matched QA pairs, $QA_p$ as the number of QA pairs in the predicted summaries, and $QA_r$ as the number of QA pairs in the reference summaries. Then we give the precision, recall, and F1 score of this evaluation metric as follows:

\begin{equation}
    precision = \frac{QA_m}{QA_p}
\end{equation}
\begin{equation}
    recall = \frac{QA_m}{QA_r}
\end{equation}
\begin{equation}
    F1 = \frac{2 * precision * recall}{precision + recall}
\end{equation}

\section{Case Study}

In this section, we will present some cases in CSDS where existing methods are prone to make mistakes.

\begin{figure*}[hbp] 
    \centering  
    \includegraphics[width=0.99\textwidth]{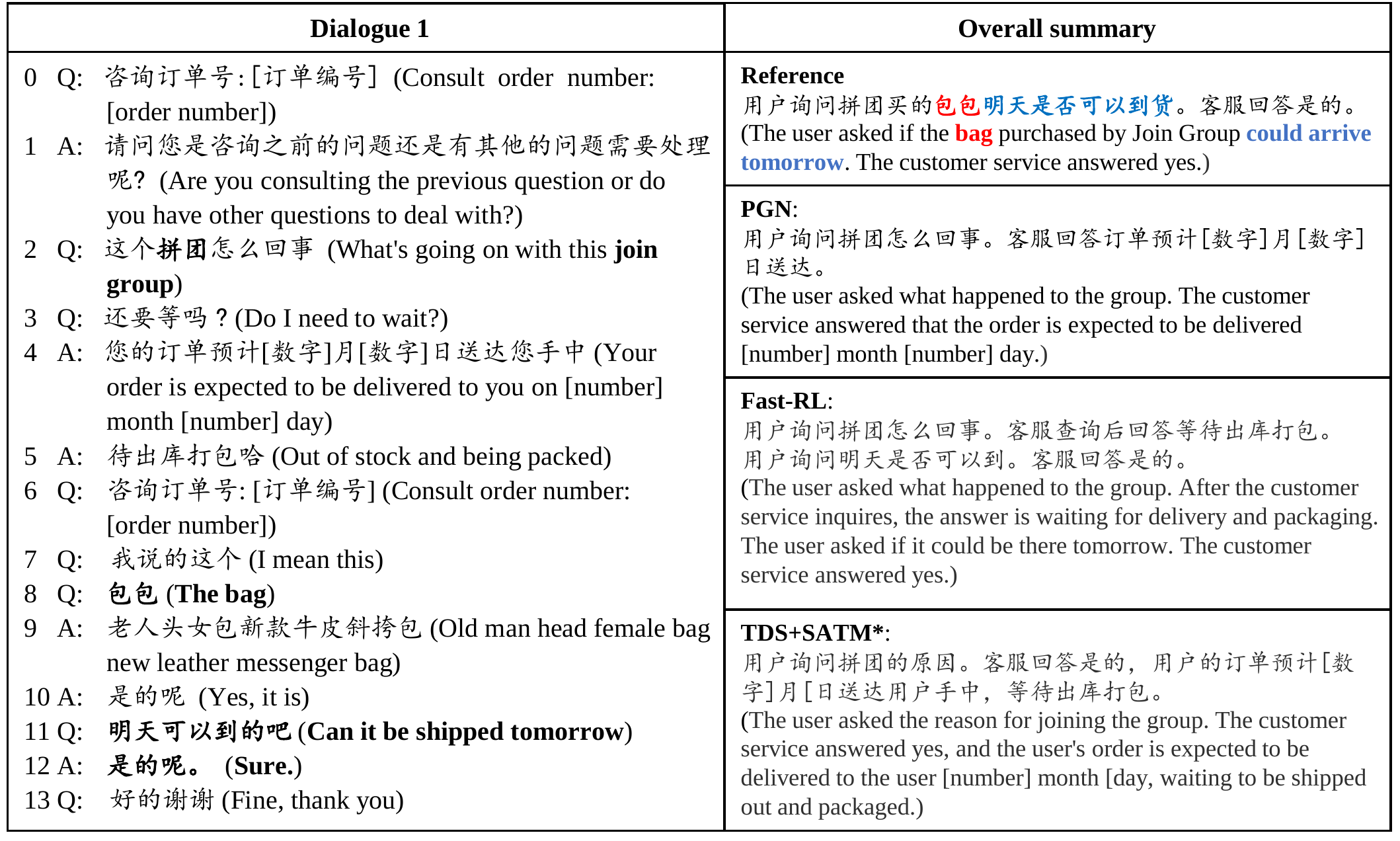}
    \caption{A real sample in CSDS test set. Some turns in the dialogue is deleted for better illustration.} 
    \label{fig:case-1}
\end{figure*}

Figure \ref{fig:case-1} shows an example where key contents are missing, and the topic structure is wrongly expressed. First, the key information ``bag'' is not summarized by any method, making the summary difficult to understand. Besides, the key question, ``could the bag arrive tomorrow'' is not extracted by most methods. Although Fast-RL correctly summarizes this question, it split the summary into two QA pairs, which actually represent the same topic and issue. Thus, it should be summarized in a single QA pair for the correct topic structure.

\begin{figure*}[hbp] 
    \centering  
    \includegraphics[width=0.99\textwidth]{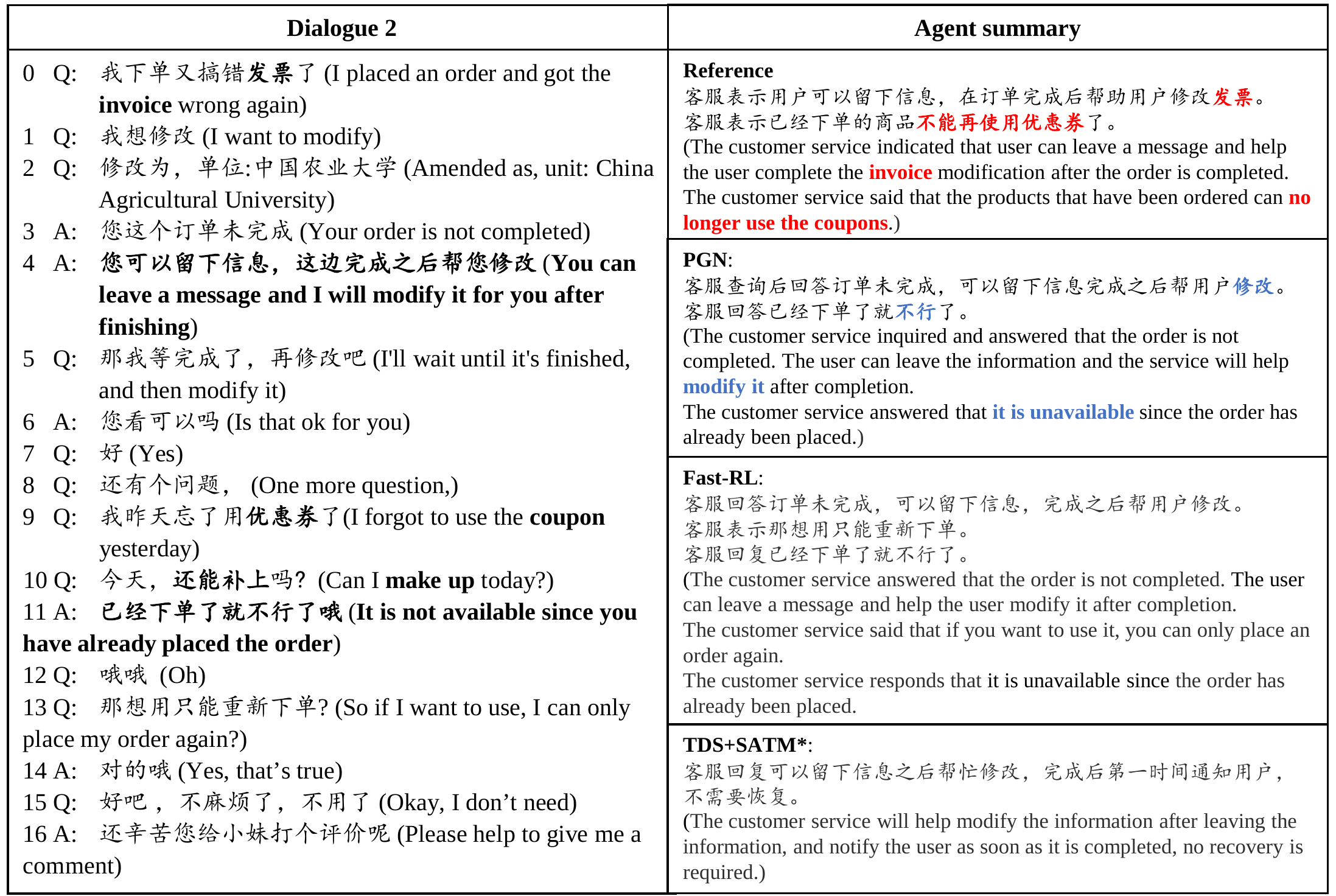}
    \caption{Another real sample in CSDS test set. Some turns in the dialogue is deleted for better illustration.} 
    \label{fig:case-2}
\end{figure*}

Figure \ref{fig:case-2} presents another example to explain the difficulty of role-oriented summary, especially for agent summary. In this case, there exist two topics, and both topics need to integrate the information from the user to obtain a complete agent summary. The key information, \textit{i.e.} ``invoice'' and ``use the coupons'', is missing in all three summarization methods. Although they can focus on correct agent utterances, they could not also focus on the necessary messages carried out by other speakers.

\end{document}